# Mini-Hes: A Parallelizable Second-order Latent Factor Analysis Model

Jialiang Wang[1, 2], Weiling Li[2*], Yurong Zhong[2], Xin Luo[1]
[1] College of Computer and Information Science, Southwest University, Chongqing 400715, China
[2] School of Computer Science and Technology, Dongguan University of Technology, Dongguan 523808, China
goallow@gmail.com, weilinglicq@outlook.com, zhongyurong91@gmail.com, luoxin21@gmail.com

*Abstract*—Interactions among large number of entities is naturally high-dimensional and incomplete (HDI) in many big data related tasks. Behavioral characteristics of users are hidden in these interactions, hence, effective representation of the HDI data is a fundamental task for understanding user behaviors. Latent factor analysis (LFA) model has proven to be effective in representing HDI data. The performance of an LFA model relies heavily on its training process, which is a non-convex optimization. It has been proven that incorporating local curvature and preprocessing gradients during its training process can lead to superior performance compared to LFA models built with first-order family methods. However, with the escalation of data volume, the feasibility of second-order algorithms encounters challenges. To address this pivotal issue, this paper proposes a mini-block diagonal hessian-free (Mini-Hes) optimization for building an LFA model. It leverages the dominant diagonal blocks in the generalized Gauss-Newton matrix based on the analysis of the Hessian matrix of LFA model and serves as an intermediary strategy bridging the gap between first-order and second-order optimization methods. Experiment results indicate that, with Mini-Hes, the LFA model outperforms several state-of-the-art models in addressing missing data estimation task on multiple real HDI datasets from recommender system. (The source code of Mini-Hes is available at https://github.com/Goallow/Mini-Hes)

## I. INTRODUCTION

High-dimensional and incomplete (HDI) data are commonly seen in big data related applications, e.g., user-item interactions in recommender system (RS). User-item interactions can be represented as a rating matrix [1], which is inherently sparse for rapidly expanding user and item scale and resource-constrained user activities. Although the rating matrix is HDI, it contains rich knowledge of entity behaviors. In order to understand them for better pattern recognition, an effective representation model is desired.

The latent factor analysis (LFA) model has proven to be effective in representing HDI data [1, 9]. It works by representing the features of each user and item with a unified latent space. Since the estimated ratings are obtained by the inner product between user's and item's latent factor vector. This leads to a fact that when updating the user's latent factor vector, it must rely on the value of the item's latent factor vector. Therefore, the LFA model is also called the bi-linear model.

To achieve the optimal prediction performance for the LFA model, there are currently three main optimization methods. First, the first-order optimization family [2, 3], e.g., vanilla gradient descent and momentum-based gradient among large descent. They acquire the update increment by scaling the gradient with the learning rate in a linear way.

The second type encompasses the Ada-Family methods [2-5]. These methods enhance the model's representative performance by preconditioning the gradient in training process to incorporate tiny curvature information. Specifically, they employ nonlinear scaling and incorporate historical gradient data to extract limited insights into the underlying curvature.

Besides, the second-order optimization family, e.g., natural gradient method [4], Hessian-free optimization [6], and quasi-Newton methods [7, 8], incorporates curvature information by computing the Hessian matrix or its variations, along with their inverses, during each iteration step to achieve enhanced generalization performance. Given the inherent high-dimensionality of the LFA model, second-order-based LFA (SLF) models typically favor Hessian-free optimization to circumvent the computational overhead associated with direct computation of the Hessian matrix and its inverse [6, 9].

However, with the growing volume of data, the cost of acquiring curvature information experiences a significant escalation. Furthermore, improving computational efficiency through parallel computing for Hessian-free-based SLF models becomes increasingly challenging, thereby presenting practical feasibility constraints.

To strike a delicate equilibrium between first-order and second-order optimization for the LFA model, i.e., predictive accuracy for missing data and computational time costs. This paper proposes a mini-block diagonal Hessian-free (Mini-Hes) optimization-based LFA model. The primary goal is to achieve a precise and efficient representation of missing data in the HDI matrix within a recommender system. Meanwhile, in pursuit of an accelerated computational process, the Mini-Hes model adopts multi-threading technology to compute each block's Hessian-vector product. The motivation behind the Mini-Hes model stems from the following two-fold principles:

(1) The Hessian matrix of the LFA model is an HDI matrix, characterized by high-dimensional as well as diagonal dominance, where the elements in the main diagonal blocks' value are larger than those in the off-diagonal blocks. As the sparsity of the input data for the LFA model increases, the Hessian matrix undergoes a concurrent expansion in dimensionality and augmentation in sparsity, leading to a progressively more pronounced predominance of the main diagonal blocks.

This work was supported in part by the National Natural Science Foundation of China under Grant 62102086, in part by the Guangdong Basic and Applied Basic Research Foundation under Grant 2022A1515140102 and Grant 2019A1515111058, and in part by the Guangdong Province Universities and College Pearl River Scholar Funded Scheme under Grant 2019.
*Corresponding author

(2) Each block operates independently, hence, computational efficiency can be improved through multi-threading concurrent computation.

With these principles, Mini-Hes model owns the ability to build an LFA model in parallel, which can accelerate the building process while maintaining the representing accuracy. Experiment results indicate that, with Mini-Hes, an LFA model outperforms several state-of-the-art models in addressing missing data estimation task on multiple real HDI datasets from recommender system.

The rest of this paper is organized as below. Section II demonstrates the preliminaries. Section III presents the Mini-Hes model. Section IV gives the experimental results. Finally, Section V draws the conclusions.

## II. PRELIMINARIES

Given a rating dataset of RS, each entry includes three essential attributes: user id ($u$), item id ($i$), and rating ($r$), organized in the form of a tuple ($u, i, r$). These entries can be described by an HDI matrix, which is defined as follows:

***Definition 1***. *(An HDI matrix)*. Given a user set $U$ and item set $I$, a matrix $\mathbf{R} \in \mathbb{R}^{|U| \times |I|}$'s element $r_{u,i}$ represents the rating between $u \in U$ and $i \in I$. Let set $\mathbf{R}_K$ and $\mathbf{R}_M$ denotes known elements and missing elements of the matrix $\mathbf{R}$. The $\mathbf{R}$ is so-called HDI matrix if and only if $|\mathbf{R}_K| \ll |\mathbf{R}_M|$.

***Definition 2***. *(An LFA model)*. Given $\mathbf{R}_K$, an LFA model is aiming to predict the missing data of $\mathbf{R}$ via constructing an estimated matrix $\hat{\mathbf{R}}$ using a matrix product of two embedding matrices, $\hat{\mathbf{R}} = \mathbf{P}^T \mathbf{Q}$, where $\mathbf{P} \in \mathbb{R}^{|U| \times f}$ and $\mathbf{Q} \in \mathbb{R}^{|I| \times f}$. Symbol $f$ represents the dimension of the latent space.

To optimize $\hat{\mathbf{R}}$, an effective metric is required for evaluating the gap between $\mathbf{R}$ and $\hat{\mathbf{R}}$. A vanilla LFA model utilizes Euclidean distance [1, 9]. The vanilla objective function of the LFA model is as follows:

$$L(\mathbf{P},\mathbf{Q}) = \frac{1}{2} \sum_{r_{u,i} \in \mathbf{R}_K} \left( \left( r_{u,i} - \sum_{d=1}^{f} p_{u,d} q_{i,d} \right)^2 + \lambda \sum_{d=1}^{f} \left( p_{u,d}^2 + q_{i,d}^2 \right) \right), \quad (1)$$

where $\lambda$ is a constant controlling the effectiveness of the Tikhonov regularization term, $p_{u,d}$ represents the element at the $u$-th row and $d$-th column of $\mathbf{P}$, $q_{i,d}$ denotes the element at the $i$-th row and $d$-th column of $\mathbf{Q}$.

## III. MINI-HES MODEL

### A. Hessian Matrix of the LFA Model

The LFA model is a so-called bi-linear model. Due to its bi-linear nature, its Hessian matrix exhibits an HDI structure with dominant diagonal blocks as shown in Fig. 1, the original Hessian-vector product. The element in the Hessian matrix of the LFA model is as follows:

(1) For the diagonal blocks, we have:

$$\begin{cases} \partial^2 L / \partial p_{u,a}^2 = \sum_{r_{u,i} \in \mathbf{R}_{Ku}} \left( q_{i,a}^2 + \lambda \right), & \partial^2 L / \partial q_{i,a}^2 = \sum_{r_{u,i} \in \mathbf{R}_{Ki}} \left( p_{u,a}^2 + \lambda \right); \\ \partial^2 L / \partial p_{u,a} \partial p_{u,b} = \sum_{r_{u,i} \in \mathbf{R}_{Ku}} q_{i,a} q_{i,b}, & \partial^2 L / \partial q_{i,a} \partial q_{i,b} = \sum_{r_{u,i} \in \mathbf{R}_{Ki}} p_{u,a} p_{u,b}; \end{cases} \quad (2)$$

where $\mathbf{R}_{Ku}$ is the subset of $\mathbf{R}_K$ w.r.t entity $u$, $p_{u,a}$ and $p_{u,b}$ denotes the $a$-th and $b$-th element of $p_u$, respectively. $\mathbf{R}_{Ki}$, $q_{i,a}$ and $q_{i,b}$ have similar definitions.

(2) Analyzing the off-diagonal blocks is more intricate than the diagonal blocks, for clarity, we categorize them into two cases:

**Case 1**: There is no interdependence among the decision variables $\mathbf{P}$ and $\mathbf{Q}$. Hence, for these blocks, their element values are zero as follows:

$$\partial^2 L / \partial p_{u_1,a} \partial p_{u_2,b} = \partial^2 L / \partial q_{i_1,a} \partial q_{i_2,b} = \partial^2 L / \partial p_{u,a} \partial q_{i,b} = 0. \quad (3)$$

Consequently, it is apparent that the values of these blocks are all equal to zero for any $u_1$ and $u_2$, as well as for $i_1$ and $i_2$. Moreover, if $u$ and $i$ do not exhibit any interaction, the blocks pertaining to them are also zero.

**Case 2**: There is an interaction between $u$ and $i$, and the values of the off-diagonal blocks related to $u$ and $i$ are as follows:

$$\begin{cases} \partial^2 L / \partial p_{u,a} q_{i,a} = p_{u,a} q_{i,a} - r_{u,i} + \sum_{d=1}^{f} p_{u,d} q_{i,d}, \\ \partial^2 L / \partial p_{u,a} \partial q_{i,b} = p_{u,b} q_{i,a}. \end{cases} \quad (4)$$

Note that the Hessian matrix of the standard LFA model has the properties of block diagonal dominant, high-dimensional, and incomplete. The size of each dominant block is $f \times f$. It indicates a dominance of the main diagonal block in the Hessian matrix, whose element values are significantly larger than those in the off-diagonal blocks, including numerous zero-value blocks in the off-diagonal blocks section.

### B. Mini-Block Diagonal Approximation

The Hessian matrix of the LFA model is characterized by properties of block diagonal dominance, high dimensionality, and sparsity. Therefore, an approximation of the LFA model's Hessian matrix can be achieved through a block-diagonal vector. The approximation function is represented as follows:

$$\mathbf{H}_L(\mathbf{P},\mathbf{Q}) \approx Diag\left( \mathbf{H}_L(\mathbf{P}_{u1}),...,\mathbf{H}_L(\mathbf{Q}_{|I|}) \right), \quad (5)$$

where $Diag()$ denotes the block-diagonal vector, $\mathbf{H}_L(\mathbf{P}_u)$ and $\mathbf{H}_L(\mathbf{Q}_i)$ is a $f$ by $f$ symmetric matrix, and denotes $k$-th block of the block-diagonal vector. After executing $Diag()$, the curse of dimensionality brought by the Hessian matrix $\mathbf{H}_L$ can be avoided, reducing the storage complexity from $O(((|U|+|I|) \times f)^2)$ to $O((|U|+|I|) \times f^2)$.

### C. Hessian-free Optimization

Currently, the second-order optimization for the learning model aims to solve the linear system as follows:

$$\arg \min_{\Delta x} \nabla L(x) + \mathbf{H}_L(x) \Delta x \leq \tau \mathbf{1}, \quad (6)$$

where $\nabla L(x)$ denotes the gradient of $L(x)$, $\Delta x$ represents an increment obtained at point $x$, $\mathbf{H}_L(x)$ corresponds to the Hessian matrix of $L(x)$, $\mathbf{1}$ represents the all-one vector, $\tau$ is the terminate condition constant. The Hessian-free optimization utilizes conjugate gradient (CG) to solve this linear system in multi-iteration in neural network models bypassing computing the Hessian inverse [6, 9]. In each iteration of CG, the Hessian-free optimization incorporates a two-fold concept: a) adopts a semi-definite Gauss-Newton matrix to approximate the non-convex objective function's Hessian matrix:

$$\mathbf{H}_L(x) \approx \mathbf{G}_L(x) = \mathbf{J}_L(x)^{\mathrm{T}} \mathbf{J}_L(x), \quad (7)$$

where $\mathbf{G}_L(x)$ denotes Gauss-Newton approximation w.r.t $\mathbf{H}_L(x)$, $\mathbf{J}_L(x)$ is the Jacobian matrix; b) utilizes $R$-operator to compute Hessian-vector as follows:

$$\mathbf{G}_L(x)v = \mathbf{J}_L(x)^{\mathrm{T}} \mathbf{J}_L(x)v = R\{\nabla L(x)\}, \quad (8)$$

where $R\{\}$ denotes the $R$-operator [6, 9, 10], $v$ is the conjugate direction in CG.

### D. Mini-block Diagonal Hessian-free Optimization

#### a. Reformulation

As the LFA model is a bi-linear model involving the product of two decision parameters, for the sake of simplifying the analysis of the Mini-Hes optimization procedure, this section will redefine the loss function of the LFA model, omitting the Tikhonov regularization term. Simultaneously, it will combine the two decision parameters, $\mathbf{P}$ and $\mathbf{Q}$, into a unified decision parameter vector denoted as $\mathbf{X} \in \mathbb{R}^{(|U|+|I|) \times f}$. The reformulated version of the LFA model's loss function is presented as follows:

$$E(\mathbf{X}) = \frac{1}{2} \sum_{r_{u,i} \in \mathbf{R}_K} \left( r_{u,i} - \sum_{d=1}^{f} x_{u,d} x_{i,d} \right)^2, \quad (9)$$

where $x_{u,d}$ denotes the element at the $u$-th row and $d$-th column of $\mathbf{X}$, the same as the element $x_{i,d}$.

#### b. Generalized Gauss-Newton Diagonal Approximation

The loss function of the LFA model is inherently non-convex, primarily due to the presence of the bi-linear term. In the case of this function, its Hessian matrix can potentially possess negative eigenvalues. As a result, a semi-definite generalized Gauss-Newton matrix is employed to approximate the Hessian matrix of equation (6). The approximation function is expressed as follows:

$$\mathbf{H}_E(\mathbf{X}) \approx Diag\left(\mathbf{G}_E(\mathbf{X}_{u1}), \ldots, \mathbf{G}_E(\mathbf{X}_{|U|}), \mathbf{G}_E(\mathbf{X}_{i1}), \ldots, \mathbf{G}_E(\mathbf{X}_{|I|})\right)$$
$$= Diag\left(\mathbf{J}_E(\mathbf{X}_{u1})^{\mathrm{T}} \mathbf{J}_E(\mathbf{X}_{u1}), \ldots, \mathbf{J}_E(\mathbf{X}_{|I|})^{\mathrm{T}} \mathbf{J}_E(\mathbf{X}_{|I|})\right), \quad (10)$$

whose $\mathbf{G}_E(\mathbf{X}_u)$ and $\mathbf{J}_E(\mathbf{X}_u)$ is the generalized Gauss-Newton matrix and Jacobian matrix of (9) for the user $u$-th block, respectively. The same as $\mathbf{G}_E(\mathbf{X}_i)$ and $\mathbf{J}_E(\mathbf{X}_i)$. According to [9], mapping the bi-linear term $x$ by $x$ into a function is helpful for executing optimization process. The mapping skill is as follows:

$$M_{(u,i)}(\mathbf{X}) = \sum_{d=1}^{f} x_{u,d} x_{i,d}, \quad (11)$$

where mapping value $M_{(u,i)}(\mathbf{X})$ is a scalar obtained by the bi-linear term. Then, the loss function (9) can be reformed as follows:

$$E(\mathbf{X}) = E(M_{(u,i)}(\mathbf{X})) = \frac{1}{2} \sum_{r_{u,i} \in \mathbf{R}_K} \left( r_{u,i} - M_{(u,i)}(\mathbf{X}) \right)^2. \quad (12)$$

Then, the Hessian-vector $\mathbf{G}_E(M_{(u,i)}(\mathbf{X}))v$ can be computed via multiple CG iteration:

$$\boldsymbol{\omega}_E(\mathbf{X}) = \mathbf{J}_E\left(M_{(u,i)}(\mathbf{X})\right)^{\mathrm{T}} \mathbf{J}_E\left(M_{(u,i)}(\mathbf{X})\right)v, \quad (13)$$

where $\mathbf{J}_E(M_{(u,i)}(\mathbf{X}))$ is the Jacobian matrix of $E(M_{(u,i)}(\mathbf{X}))$, and $v$ represents the conjugate direction in each CG epoch. And the Hessian-vector product $\boldsymbol{\omega}_E(\mathbf{X})$ can be computed via two times matrix-vector product. For the term involving the Jacobian matrix-vector product $\mathbf{J}_E(M_{(u,i)}(\mathbf{X}))v$, $R$-operator efficiently computes it as follow:

$$\mathbf{J}_E\left(M_{(u,i)}(\mathbf{X})\right)v = \left( R\left(M_{(u,i)}(\mathbf{X})\right)\bigg|_{(u,i) \in \mathbf{R}_K} \right). \quad (14)$$

For the block's level, (14) can be expanded as follows:

$$\begin{cases} \text{For each user block, } \forall u \in U, \, d = 1 \sim f: \\ \mathbf{J}_E\left(M_{(u,i)}(\mathbf{X}_u)\right)v_u = \left( \sum_{d=1}^{f} \left( v_{u,d} x_{i,d} \right) \bigg|_{(u,i) \in \mathbf{R}_{Ku}} \right), \\ \text{For each item block, } \forall i \in I, \, d = 1 \sim f: \\ \mathbf{J}_E\left(M_{(u,i)}(\mathbf{X}_i)\right)v_i = \left( \sum_{d=1}^{f} \left( x_{u,d} v_{i,d} \right) \bigg|_{(u,i) \in \mathbf{R}_{Ki}} \right), \end{cases} \quad (15)$$

The rest Jacobian matrix $\mathbf{J}_E(M_{(u,i)}(\mathbf{X}))^{\mathrm{T}}$ can be calculated as follows:

$$\mathbf{J}_E\left(M_{(u,i)}(\mathbf{X})\right)^{\mathrm{T}} = \left( \partial M_{(u,i)}(\mathbf{X}) \big/ \partial \mathbf{X} \bigg|_{(u,i) \in \mathbf{R}_K} \right)^{\mathrm{T}}. \quad (16)$$

Combining (15) and (16), we can have the approximation of Gauss-Newton vector product for each user block and item block:

$$\begin{cases} \text{For each user block, } \forall u \in U, \, d = 1 \sim f: \\ \boldsymbol{\omega}_E(\mathbf{X})_{u,d} = \sum_{i \in \mathbf{R}_{Ku}} \left( x_{i,d} \left( \sum_{d=1}^{f} \left( v_{u,d} x_{i,d} \right) \right) \right), \\ \text{For each item block, } \forall i \in I, \, d = 1 \sim f: \\ \boldsymbol{\omega}_E(\mathbf{X})_{i,d} = \sum_{u \in \mathbf{R}_{Ki}} \left( x_{u,d} \left( \sum_{d=1}^{f} \left( x_{u,d} v_{i,d} \right) \right) \right). \end{cases} \quad (17)$$

where $\mathbf{R}_{Ku}$ and $\mathbf{R}_{Ki}$ respectively denotes the subsets of RK related to user $u$ and item $i$, $\boldsymbol{\omega}_E(\mathbf{X})_{u,d}$ and $\boldsymbol{\omega}_E(\mathbf{X})_{i,d}$ respectively denote the $d$-th element of sub-vector in $\boldsymbol{\omega}_E(\mathbf{X})$ associated with user u and item $i$.

#### c. Incorporating Tikhonov Regularization Term

With the incorporation of the Tikhonov regularization term, the objective function is as follows:

$$L(\mathbf{X}) = \frac{1}{2} \sum_{r_{u,i} \in \mathbf{R}_K} \left( \left( r_{u,i} - \sum_{d=1}^{f} x_{u,d} x_{i,d} \right)^2 + \lambda \sum_{d=1}^{f} \left( x_{u,d}^2 + x_{i,d}^2 \right) \right). \quad (18)$$

The Hessian-vector $\boldsymbol{\omega}_T(\mathbf{X})$ for the Tikhonov regularization term is obtained by the subtraction of $\boldsymbol{\omega}_E(\mathbf{X})$ from $\boldsymbol{\omega}_L(\mathbf{X})$

$$\boldsymbol{\omega}_T(\mathbf{X}) = \boldsymbol{\omega}_L(\mathbf{X}) - \boldsymbol{\omega}_E(\mathbf{X}) = \mathbf{H}_L(\mathbf{X})v - \mathbf{H}_E(\mathbf{X})v. \quad (19)$$

Incorporating a damping term $\gamma$ into the Gauss-Newton step $\mathbf{G}_L(\mathbf{X})$ can be viewed as an augmentation of the curvature matrix with a Tikhonov regularization component, thereby promoting positivity and enhancing the stability of the Hessian approximation. The function is expressed as follows:

$$\omega_L(\mathbf{X}) = (\mathbf{G}_L(\mathbf{X}) + \gamma \mathbf{I})$$
$$\approx \mathbf{G}_E(\mathbf{X})v + \gamma v + (\mathbf{H}_L(\mathbf{X})v - \mathbf{H}_E(\mathbf{X})v)$$
$$\Rightarrow \begin{cases} \text{For each user block, } \forall u \in U, d = 1 \sim f: \\ \omega_L(\mathbf{X})_{u,d} = \sum_{i \in R_{Ku}} \left( x_{i,d} \left( \sum_{d=1}^{f} (v_{u,d} x_{i,d}) \right) \right) + \gamma v_{u,d} + \lambda v_{u,d} |R_{Ku}|, \\ \text{For each item block, } \forall i \in I, d = 1 \sim f: \\ \omega_L(\mathbf{X})_{u,d} = \sum_{u \in R_{Ki}} \left( x_{u,d} \left( \sum_{d=1}^{f} (x_{u,d} v_{i,d}) \right) \right) + \gamma v_{i,d} + \lambda v_{i,d} |R_{Ki}|. \end{cases}$$
(20)

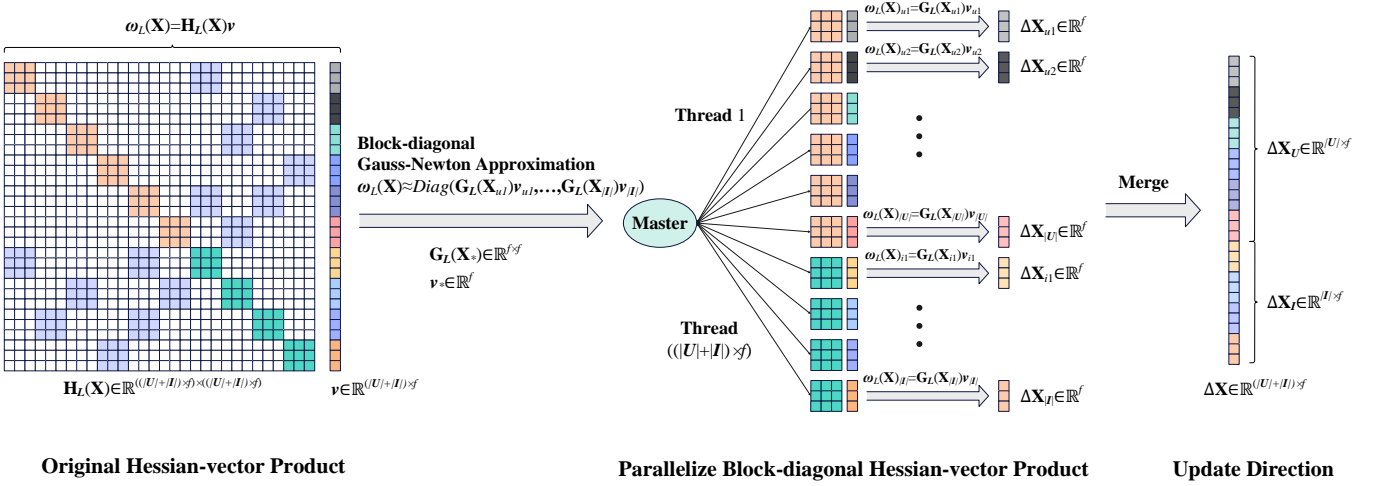

Fig. 1. Schematic of Mini-Hes Multithreaded Concurrent Computation

*d. Iterative Update Rule*

After multiple CG iterations, we can derive the increment $\Delta \mathbf{X}$ for updating variable $\mathbf{X}$ at $t$-th epoch:

$$\mathbf{X}^{t+1} = \mathbf{X}^t + \Delta \mathbf{X}^t. \quad (21)$$

*E. Block Concurrent Computation*

For each block's Hessian-vector product

$$\mathbf{H}_L(\mathbf{X})v \approx Diag\left(\mathbf{G}_L(\mathbf{X}_{u1})v_{u1}, ..., \mathbf{G}_L(\mathbf{X}_{|I|})v_{|I|}\right). \quad (22)$$

As shown in Fig. 1, since each Hessian matrix block for every user or item is independent of each other, the sub-component of $v$ for different users or items is also independent. It is possible to concurrently compute the Hessian-vector products for each block and component of $v$ on different threads. Stated differently, each individual block has the capability to concurrently multiply its respective subset of $v$ without introducing interference to the ultimate outcome. Hence, within this paper, employing a Master-workers multi-threading computation strategy facilitates efficient utilization of processor resources, thereby reducing the computational time costs of the algorithm. Each block's Hessian-vector operation, associated with elements of $v$, is allocated to an individual thread for execution. The adjustment of thread count offers control over the simultaneous execution of block-specific Hessian-vector operations related to elements of $v$.

## IV. EXPERIMENT

In this section, experiments have been designed to verify the performance of the Mini-Hes model. All experiments are conducted on a Linux server with Ubuntu 20.04.6 LTS 64bit, two-way Intel Xeon Silver 4214R 2.4 GHz, 12 cores and 24 threads for each processor, and 128 GB RAM. Moreover, all tested models are executing on OpenJDK 11 LTS.

*A. General Settings*

**Datasets**. Four public HDI datasets, e.g., Flixster [11], Rating Disposition (RD) [12], Movie-Lens 10M (ML-10M) [13] and MovieLens-20M (ML-20M) [13], are utilized in this work,. Their details are summarized in Table II.

TABLE I.  INVOLVED DATASETS

| No. | Name | $|U|$ | $|I|$ | $R_K$ | Density |
|---|---|---|---|---|---|
| D1 | Flixster | 147,612 | 48,794 | 8,196,077 | 0.11% |
| D2 | RD | 6,724 | 68,044 | 3,908,657 | 0.85% |
| D3 | ML-10M | 71,567 | 10,681 | 10,000,054 | 1.31% |
| D4 | ML-20M | 138493 | 27278 | 20,000,263 | 0.53% |

All datasets are randomly divided into three parts, namely training set, testing set, and validation set with partition ratio 6:2:2.

**Evaluation metrics**. The Root Mean Square Error (RMSE) and Mean Absolute Error (MAE) are adopted to evaluate performance of involved LFA models by transforming representation task into missing data prediction task [1, 9]. Let $\Lambda$ denotes the dataset for evaluation, RMSE and MAE can be derived as follows:

$$\text{RMSE} = \sqrt{\frac{\sum_{r_{u,i}} \left( r_{u,i} - \sum_{d=1}^{f} x_{u,d} x_{i,d} \right)^2}{|\Lambda|}},$$
$$\text{MAE} = \frac{\sum_{r_{u,i}} \left| r_{u,i} - \sum_{d=1}^{f} x_{u,d} x_{i,d} \right|_{abs}}{|\Lambda|}, \quad (23)$$

where $|\Lambda|$ is the volume of $\Lambda$, and $|\ |_{abs}$ is the operator for calculating absolute value.

*B. Performance Benchmark*

**Baselines**. Four state-of-the-art methods are chosen as benchmarks. Their details are given as follows:

**M1. Adam-based LFA model** [14]: It is a canonical adaptive gradient descent optimizer widely adopted in many model's training process. The Adam-based LFA model can achieve high representation performance by rescaling the gradient via the first-order and the second-order momentum.

**M2. YOGI-based LFA model** [15]: In certain scenarios, adaptive gradient optimizers struggle to drive the training model towards a stable convergence. To address this issue, the YOGI optimizer implements controlled management of the effective learning rate increment, resulting in enhanced performance.

**M3. ACProp-based LFA Model** [16]: It is an adaptive optimizer which incorporates centering of second-order momentum and asynchronous update process outperforming gradient descent and other adaptive optimizers.

**M4. SLF model** [9]: It is a classical second-order LFA model which optimized by the Hessian-free optimization. Since the assimilation of the curvature information, the Hessian-free optimization methods can outperform other first-order optimizers.

**M5. Mini-Hes model**: The model proposed in this work.

**Training settings**. All models' setting strategies are given as follows:

(1) All involved models' initial latent factor matrices are randomly sampling from U(0,0.04). According to [9, 14-16], each model's hyperparameters recommended setting, e.g., learning rate, damping term, weighting parameters $\beta_1$ and $\beta_2$, are fine-tunning via grid search on validation set. Except for the latent dimension $f$, and the regularization term's constant $\lambda$, all comparable models' hyperparameters search range and search gap are set to theirs recommended value. In this section, $f$ is fixed as 20, and the search range of $\lambda$ is [0.0, 0.1] as well as the gap is set to 0.01. The difference of hyperparameter between M4 and M5 is that M4's $\tau$ is 100, but M5's $\tau$ will be selected as 0.1 or 1.

(2) The same terminate condition is adopted for each model's training process. In details, the maximum training upper bound is set to 500 epochs, and all models take a same early stop strategy, i.e., if the validate error bigger than the lowest point over 10 epochs, then record the best fine-tunning model's parameter and test its generalization error on test set.

*a. Multi-threaded Computation Efficiency Analysis*

TABLE II. SPEEDUP BENCHMARK FOR M5 ON D1-D4

| Dataset | Thread | Time (Sec) | Speedup |
|---|---|---|---|
| D1 | 2 | 137.272±4.485 | 1 |
|  | 4 | 71.242±4.605 | 1.93 |
|  | 8 | 45.236±2.989 | 3.03 |
|  | 16 | 26.535±1.372 | 5.17 |
|  | 32 | 25.588±1.237 | 5.36 |
| D2 | 2 | 51.285±0.084 | 1 |
|  | 4 | 29.478±1.058 | 1.74 |
|  | 8 | 19.561±1.713 | 2.62 |
|  | 16 | 13.998±0.182 | 3.66 |
|  | 32 | 11.311±0.291 | 4.53 |
| D3 | 2 | 148.775±0.449 | 1 |
|  | 4 | 84.719±9.396 | 1.75 |
|  | 8 | 47.190±2.277 | 3.15 |
|  | 16 | 26.168±1.155 | 5.69 |
|  | 32 | 18.065±1.066 | 8.23 |
| D4 | 2 | 378.711±7.069 | 1 |
|  | 4 | 201.622±2.186 | 1.88 |
|  | 8 | 120.144±1.157 | 3.15 |
|  | 16 | 70.242±1.259 | 5.39 |
|  | 32 | 44.233±1.588 | 8.56 |

This section primarily presents the empirical performance of M5 in computing MAE on datasets D1 to D4, utilizing various thread configurations, 2-thread to 32-thread. In this experimental context, the 2-thread configuration serves as the reference point. Threads are equitably distributed, with half of them dedicated to computations involving the user latent feature matrix, while the remaining half focuses on computations related to the item latent feature matrix.

According to Table II, on D1 to D4, for the same dataset, a higher number of threads results in a larger speedup ratio. For example, D4 is the largest dataset among the benchmark datasets in this study. When tested with two threads, the average runtime for model M5 was 378.711 seconds. However, when using four threads, the runtime reduced to 201.622 seconds, resulting in a speedup ratio of 1.88, which nearly exhibits linear scaling. As the thread count increased to eight, the execution time further decreased to 120.144 seconds, achieving a speedup ratio of 3.15. Further increasing the thread count to 16 resulted in an execution time of 70.242 seconds and a speedup ratio of 5.39. When the thread count was increased to 32, the runtime further decreased to just 44.233 seconds, yielding a remarkable speedup ratio of 8.56.

*b. Comparison against the state-of-the-art methods*

(1) **When compared to a full Gauss-Newton matrix approximation, Mini-Hes maintains adequate second-order information and shows considerable accuracy in representation tasks**. For example, on D1, M4 and M5 achieved similar prediction results using RMSE. Specifically, M4 had average RMSE values of 0.8505, while M5's results were 0.8508. According to the Tables III-VI, there are similar results in benchmarks on other datasets D1 to D4, especially with regards to the RMSE metric. However, the situation is different when it comes to the MAE metric. M5's MAE values are 0.68% lower on D1, 0.54% lower on D2, 0.41% lower on D3, 0.39% lower on D4 compared to M4.

(2) **The block-diagonalized approximation makes a Hessian-free-based second-order LFA model parallelizable on multiple threads**. As shown in Tables III-VI, it is evident that employing the strategy of block-diagonal concurrent computation results in a significant reduction in time costs. Taking the time cost results under the RMSE metric as an example, M5 incurs average time costs of 27.763 seconds, 13.034 seconds, 20.010 seconds, and 45.762 seconds on datasets D1 to D4, respectively. In contrast, M4 exhibits significantly higher average time costs, with average time expenditures of 990.824 seconds, 248.976 seconds, and 1131.051 seconds, and 4092.897 seconds on D1 to D4. To put it differently, the average time costs of M5 on datasets D1 to D4 are only 2.80%, 5.24%, 1.77%, and 1.12% of those incurred by M4. Similar results are also observed under the MAE metric.

(3) **Surprisingly the incorporate of the curvature information for the LFA model, the Mini-Hes model can achieve higher prediction accuracy compared with others adaptive gradient based LFA models**. On datasets D1 to D4, M5 outperformed all competitors, including M1 to M3, in terms of MAE values. Regarding the RMSE metric, M5 also outperformed all the adaptive gradient based LFA models, M1 to M3, except for the D2 dataset. For example, on the D4 dataset, M5 achieved lower RMSE values compared to M1, M2, and M3 by 0.05%, 0.45%, and 0.14%. Additionally, M5's

MAE value is 0.62% lower than M1, 0.95% lower than M2, and 0.60% lower than M3 on the same dataset.

TABLE III. PERFORMANCE BENCHMARK ON D1

| Model | Metric | Time (Sec) | Epoch |
|---|---|---|---|
| M1 | RMSE 0.8531 ±0.0007 | 778.858 ±10.101 | 437 ±1 |
| | MAE 0.6301 ±0.0008 | 86.697 ±2.679 | 47 ±0 |
| M2 | RMSE 0.8513 ±0.0008 | 147.999 ±1.647 | 62 ±0 |
| | MAE 0.6268 ±0.0006 | 150.683 ±0.718 | 64 ±0 |
| M3 | RMSE 0.8518 ±0.0007 | 765.135 ±91.949 | 374 ±57 |
| | MAE 0.6270 ±0.0005 | 617.551 ±36.699 | 299 ±0 |
| M4 | RMSE **0.8505 ±0.0005** | 990.824 ±33.715 | **23 ±0** |
| | MAE 0.6270 ±0.0002 | 788.698 ±24.787 | **19 ±0** |
| M5 | RMSE 0.8508 ±0.0006 | **27.763 ±1.534** | 36 ±0 |
| | MAE **0.6227 ±0.0003** | 25.588 ±1.237 | 32 ±0 |

TABLE IV. PERFORMANCE BENCHMARK ON D2

| Model | Metric | Time (Sec) | Epoch |
|---|---|---|---|
| M1 | RMSE 0.7913 ±0.0011 | 361.278 ±14.112 | 421 ±1 |
| | MAE 0.5776 ±0.0005 | 246.988 ±11.827 | 288 ±2 |
| M2 | RMSE 0.7920 ±0.0012 | 79.216 ±3.388 | 68 ±0 |
| | MAE 0.5778 ±0.0005 | 78.811 ±0.356 | 70 ±0 |
| M3 | RMSE **0.7896 ±0.0007** | 270.425 ±72.165 | 286 ±71 |
| | MAE 0.5745 ±0.0002 | 229.186 ±0.547 | 247 ±1 |
| M4 | RMSE 0.7910 ±0.0013 | 248.976 ±23.390 | 36 ±0 |
| | MAE 0.5763 ±0.0005 | 221.992 ±36.099 | **25 ±5** |
| M5 | RMSE 0.7907 ±0.0014 | **13.034 ±1.597** | **35 ±1** |
| | MAE **0.5732 ±0.0005** | 11.311 ±0.291 | 32 ±0 |

TABLE V. PERFORMANCE BENCHMARK ON D3

| Model | Metric | Time (Sec) | Epoch |
|---|---|---|---|
| M1 | RMSE 0.7900 ±0.0007 | 699.288 ±0.634 | 351 ±1 |
| | MAE 0.6074 ±0.0008 | 752.516 ±46.405 | 361 ±1 |
| M2 | RMSE 0.7934 ±0.0002 | 201.454 ±38.641 | 74 ±13 |
| | MAE 0.6092 ±0.0005 | 178.387 ±1.826 | 67 ±1 |
| M3 | RMSE 0.7905 ±0.0004 | 676.194 ±17.770 | 304 ±1 |
| | MAE 0.6076 ±0.0005 | 636.304 ±93.117 | 281 ±43 |
| M4 | RMSE **0.7890 ±0.0005** | 1131.051 ±33.450 | 101 ±1 |
| | MAE 0.6063 ±0.0007 | 1019.950 ±276.704 | 90 ±20 |
| M5 | RMSE 0.7897 ±0.0005 | **20.010 ±0.772** | **32 ±0** |
| | MAE **0.6038 ±0.0006** | 18.065 ±1.066 | **18 ±1** |

TABLE VI. PERFORMANCE BENCHMARK ON D4

| Model | Metric | Time (Sec) | Epoch |
|---|---|---|---|
| M1 | RMSE 0.7821 ±0.0005 | 1460.779 ±5.079 | 357 ±1 |
| | MAE 0.5967 ±0.0002 | 1291.316 ±83.233 | 300 ±1 |
| M2 | RMSE 0.7852 ±0.0003 | 434.265 ±110.855 | 75 ±15 |
| | MAE 0.5987 ±0.0001 | 376.965 ±17.654 | 66 ±1 |
| M3 | RMSE 0.7828 ±0.0002 | 1920.911 ±6.695 | 384 ±1 |
| | MAE 0.5966 ±0.0001 | 1184.629 ±74.112 | 249 ±1 |
| M4 | RMSE **0.7808 ±0.0005** | 4092.897 ±13.694 | 105 ±0 |
| | MAE 0.5953 ±0.0003 | 3237.317 ±189.199 | 65 ±17 |
| M5 | RMSE 0.7817 ±0.0004 | **45.726 ±0.495** | **38 ±0** |
| | MAE **0.5930 ±0.0003** | 44.233 ±1.588 | 44 ±2 |

## V. CONCLUSION

The second-order LFA model can outperform the first-order optimization-based LFA model in representation accuracy. However, as the volume of the data increases, the decision parameter vectors of the LFA model can be very huge. Although the Hessian-free-based LFA model does not require the computation of the Hessian matrix and its inverse, the cost of multiple conjugate gradient iterations is not feasible for massive data scenarios. In this work, based on the discovery that the Hessian matrix of the LFA model is the HDI matrix and that the main block-diagonal vector dominate, we propose Mini-Hes which approximates the Hessian matrix using the block-diagonal vector of the LFA model, and perform Hessian-free optimization on multiple blocks concurrently by multi-threading technology, to further improve computational efficiency. Experiment results indicate that the Mini-Hes model outperforms several state-of-the-art models in addressing the task of missing data estimation on multiple HDI datasets from recommender system. Mini-Hes is an open framework for second-order optimization analysis, specifically tailored for high-dimensional, sparse, and incomplete data scenarios. Not confined solely to Hessian-free optimization, our future plans to involve assessing the performance of other second-order optimization algorithms within Mini-Hes, exploring its representation capabilities in tensor scenarios, and providing convergence analyses.